\documentclass[conference]{IEEEtran}
\IEEEoverridecommandlockouts
\usepackage{cite}
\usepackage{amsmath,amssymb,amsfonts}
\usepackage{algorithmic}
\usepackage{graphicx}
\usepackage{hyperref}
\usepackage{multirow}
\usepackage{textcomp}
\usepackage{xcolor}
\def\BibTeX{{\rm B\kern-.05em{\sc i\kern-.025em b}\kern-.08em
    T\kern-.1667em\lower.7ex\hbox{E}\kern-.125emX}}
\makeatletter
 \let\old@ps@headings\ps@headings
 \let\old@ps@IEEEtitlepagestyle\ps@IEEEtitlepagestyle
 \def\confheader#1{%
 \def\ps@IEEEtitlepagestyle{%
 \old@ps@IEEEtitlepagestyle%
 \def\@oddhead{\strut\hfill#1\hfill\strut}%
 \def\@evenhead{\strut\hfill#1\hfill\strut}%
 }%
 \ps@headings%
 }
 \makeatother

\confheader{%
 2024 International Conference on Intelligent Algorithms for Computational Intelligence Systems (IACIS)
 }

 \usepackage[pscoord]{eso-pic}
\newcommand{\placetextbox}[3]{
 \setbox0=\hbox{#3}
 \AddToShipoutPictureFG*{ \put(\LenToUnit{#1\paperwidth},\LenToUnit{#2\paperheight}){\vtop{{\null}\makebox[0pt][c]{#3}}}
 }
 }
 \placetextbox{.23}{0.055}{\small{979-8-3503-6066-0/24/\$31.00~\copyright2024 IEEE}}

\usepackage{cite}
\usepackage{amsmath,amssymb,amsfonts}
\usepackage{algorithmic}
\usepackage{graphicx}
\usepackage{textcomp}
\def\BibTeX{{\rm B\kern-.05em{\sc i\kern-.025em b}\kern-.08em
    T\kern-.1667em\lower.7ex\hbox{E}\kern-.125emX}}

\begin{document}

\title{Generation of Indian Sign Language Letters, Numbers, and Words\\
{\footnotesize \textsuperscript{}}

}

\author{
    \IEEEauthorblockN{Ajeet Kumar Yadav}
    \IEEEauthorblockA{
        \textit{Electrical Communication Engineering} \\
        \textit{Indian Institute of Science, Banglore}\\
        Bengaluru, India \\
        ajeety@iisc.ac.in
    }
    \and
    \IEEEauthorblockN{Nishant Kumar}
    \IEEEauthorblockA{
        \textit{Computer Science and Automation} \\
        \textit{Indian Institute of Science, Banglore}\\
        Bengaluru, India \\
        nishantk@iisc.ac.in
    }
    \and
    \IEEEauthorblockN{Rathna G N}
    \IEEEauthorblockA{
        \textit{Electrical Engineering} \\
        \textit{Indian Institute of Science, Banglore}\\
        Bengaluru, India \\
        rathna@iisc.ac.in
    }
}

\maketitle

\begin{abstract}
Sign language, which contains hand movements, facial expressions and bodily gestures, is a significant medium for communicating with hard-of-hearing people. A well-trained sign language community communicates easily, but those who don't know sign language face significant challenges. Recognition and generation are basic communication methods between hearing and hard-of-hearing individuals. Despite progress in recognition, sign language generation still needs to be explored. The Progressive Growing of Generative Adversarial Network (ProGAN) excels at producing high-quality images, while the Self-Attention Generative Adversarial Network (SAGAN) generates feature-rich images at medium resolutions. Balancing resolution and detail is crucial for sign language image generation. We are developing a Generative Adversarial Network (GAN) variant that combines both models to generate feature-rich, high-resolution, and class-conditional sign language images. Our modified Attention-based model generates high-quality images of Indian Sign Language letters, numbers, and words, outperforming the traditional ProGAN in Inception Score (IS) and Fréchet Inception Distance (FID), with improvements of 3.2 and 30.12, respectively. Additionally, we are publishing a large dataset incorporating high-quality images of Indian Sign Language alphabets, numbers, and 129 words.
\end{abstract}


\section{Introduction}
 According to the United Nations (UN) 2023, more than 300 sign languages are used by hard-of-hearing people worldwide. Like spoken languages with regional variations, sign languages often have distinct forms across countries. This Visual sign language is complex and involves grammatical and punctuation structures. In recent years, researchers have actively worked on sign languages like American Sign Language and British Sign Language. However, Indian Sign Language (ISL) is still far from data-driven tasks like machine translation, recognition, and generation. Extensive research on sign language recognition has been conducted for various sign languages worldwide\cite{b2}, including ISL\cite{b4}. Sign language generation is a much-underrated task hardly tried by a few people \cite{b3}, and the ISL generation is almost untouched using the modern image and video generation neural networks. The recent strides in Language Processing such as Text Classification\cite{b7}, Translation\cite{b8}, and Generation\cite{b36} present valuable opportunities for creating tools that facilitate improved communication with individuals who are hard of hearing. However, sign languages still lack sufficient high-quality labelled data for training large neural networks due to the limited availability of certified interpreters. According to the Indian Sign Language Research and Training Center (ISLRTC), only 300 certified sign language interpreters exist in India. 
    The study by Ho, Jonathan et al. \cite{b28} claims that the quality of text-conditioned video generation can be improved through joint training on both image and video datasets. While end-to-end sign language video generative models are still underdeveloped, progress can be made by first building a sign language image generative model. Based on this idea, our work aims to generate images of ISL letters, numbers, and sentences by splitting them into their constituent words using a model trained on an image dataset. We are introducing an improved image generative architecture combining a Self-attention mechanism\cite{b10} and a ProGAN\cite{b11} with state-of-art GAN loss function Wasserstein GAN with Gradient Penalty (WGAN-GP)\cite{b12}.
    The generator outputs must be exceptionally clear, with precise finger articulation and structure, to ensure accurate sign recognition in the given image. We are developing this model with progressively growing convolutional layers to achieve high-quality images. Additionally, we are incorporating self-attention layers into the architecture to enhance clarity and detail by focusing on image specifics.
    This model is tested on two datasets mentioned in section (III.A). Our other contribution is to create a dataset containing high-quality images of ISL letters, numbers, and words available publicly \href{https://github.com/rathnagnr/Generation-of-Indian-Sign-Language-Letters-and-Numbers}{here}. Our model gives a 37.32 Inception score on the Gunji, Bala, et al. dataset\cite{b32}, which is 2.47 more compared to ProGAN. Because of progressively growing layers in multiple stages, it can generate images of many resolutions, significantly reducing its training and inference time complexity and resources. This model includes two self-attention layers between two different resolution stages, focusing on the connectivity between all important regions of the images. These self-attention layers ensure that the model captures and preserves essential details. The weights learned by this model on an image dataset can be leveraged for video generation after suitable modifications in the model. Using the outcomes of this baseline model, sign language video clips can be generated by the recently suggested Frame Interpolation methods\cite{b31}.

\section{Related Work}
    \textbf{Sign Language Recognition:} 
    Computer vision researchers have focused on sign language recognition for over three decades, particularly Isolated Sign Language Recognition (ISLR) \cite{b18}\cite{b6}. More recently, there has been a shift towards addressing the more challenging task of Continuous Sign Language Recognition (CSLR) \cite{b17}. However, a significant portion of the research relies heavily on manual feature representations and statistics.
    \begin{figure}[t]
        \centering
        \includegraphics[width=0.48\textwidth, height = 0.46\textwidth]{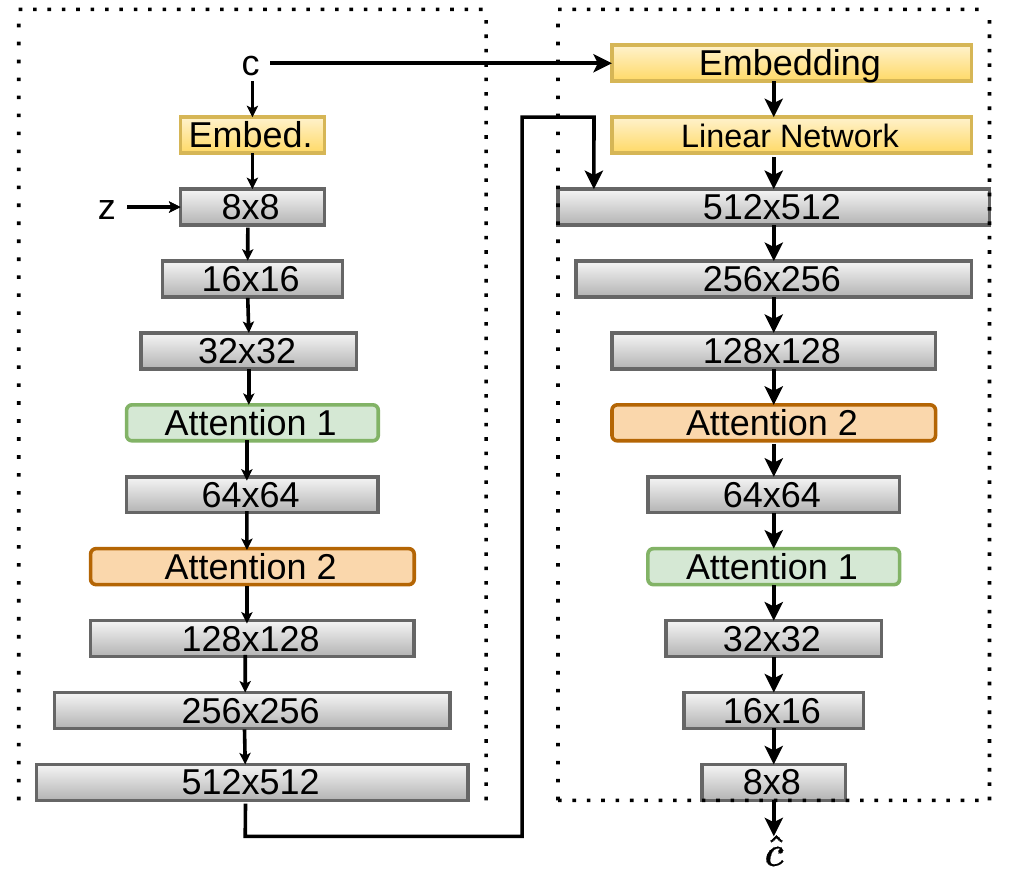}
        \caption{This is the complete architecture of our model. The left structure in the figure is the generator network, which embeds class labels with noise and starts with a resolution stage of 8x8 instead of 4x4. Two Self-attention layers are added at 64x64 and 128x128 resolution stages. The right structure in the figure is the discriminator network, which takes output from the generator while training. The discriminator embeds class labels followed by a linear network and concatenates them with the generated image. It also has two attention layers at 128x128 and 64x64 resolution stages.}
        \label{Figure 1}
    \end{figure}
    With the advent of larger datasets, there has been a notable transition to incorporating deep learning approaches, including Convolutional Neural Networks and Recurrent Neural Networks. Recently, Sign Language translation (SLT) has emerged as a task introduced by Camgoz et al.\cite{b18} to translate sign videos directly into spoken language sentences. In our project following Camgoz et al. to evaluate the Bilingual Evaluation Understudy (BLEU) score\cite{b37}, we are converting the generated image into text by pre-trained InceptionV3 model\cite{b20}.
    
    \textbf{Sign Language Production:} Initial methodologies in Sign Language Production (SLP) were based on some animated avatars\cite{b25}, capable of producing realistic, authentic sign expressions. However, these approaches heavily rely on phrase lookup and pre-generated sequences, and creating motion capture data is costly. In 2018, Stoll et al.\cite{b29} introduced an initial SLP model employing a combination of Neural Machine Translation and GAN\cite{b21}. The authors adopt an approach wherein the problem is divided into three distinct processes, which results in generating a concatenation of isolated 2D skeleton poses derived from sign glosses through a lookup table, followed by GAN to convert the skeleton into a human-like structure. Ben Saunders et al.\cite{b3} focused on automatic sign production and learning the mapping between text and skeleton pose sequences using Progressive transformers. Progressive transformers are modified for translation tasks; they receive text and progressively generate the corresponding 3D poses. The progressive transformer has a counter-embedding layer, serving as a temporal embedding. This layer furnishes the model with essential information regarding the length and speed of each sign pose sequence, which is crucial in determining the duration of the sign. This mechanism enables continuous sequence generation during training and inference.
    
    \textbf{Generative Adversarial Networks:} Image and video generation has witnessed diverse approaches, leveraging neural network architectures propelled by advancements in deep learning. Although they are image-generative models, they can be the backbone of sign language image generation. In their 2014 paper\cite{b21}, Ian J. Goodfellow et al. introduced a novel image-generative model. This model is capable of generating real-world colourful images without any class condition. Keeping this in mind, Mirza and Osindero\cite{b22} developed Conditional Generative Adversarial Nets by involving conditional information in both the Generator and Discriminator. For high-quality images, Karras, Tero, et al. proposed Progressive Growing GANs\cite{b11}. This architecture has many convolutional layers whose resolution grows progressively in several stages. The network is trained to generate and discriminate images at a specific resolution during each stage. After training the generator and discriminator at a resolution stage, it moves to the next stage with a higher resolution. This training method in multiple stages helps stabilise the training process and prevents issues like mode collapse.

    Self-attention\cite{b10}, which allows the model to capture long-range dependencies within the data, is also introduced in Self-Attention Generative Adversarial Networks\cite{b24}. Attention maps highlight the important regions in the images; as a result, the model learns global structures. These global structures play a vital role in generating high-quality and information-rich images at low and high resolution.

\section{Sign language image generation}
    Sign language contains letters, numbers, words, phrases and sentences in the image and video format. For accurate sign language recognition, the generated images require smooth and clear visibility of fingers, faces, and body posture. To facilitate real-time applications, the generative model must be robust in generating high-quality images with clear finger articulation across all sign variations. This level of detail should allow for swift and confident decision-making during the recognition process. The ProGAN generative model produces high-quality images but struggles with fine details and clarity in complex images. On the other hand, SAGAN excels at capturing essential nuances at medium resolutions (64x64 and 128x128). To achieve an ideal balance between high resolution and nuanced detail, we
    \begin{figure}[t]
        \centering
        \includegraphics[width=.49\textwidth, height = 0.23\textwidth]{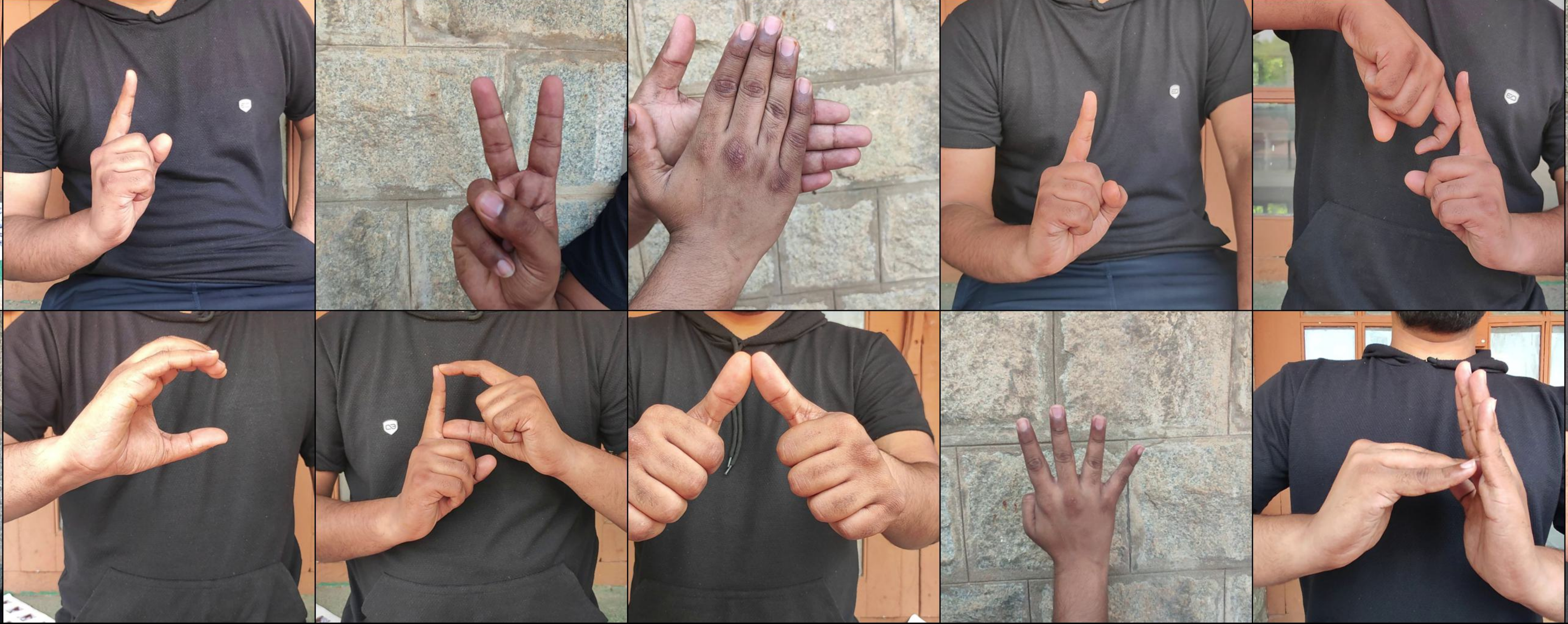}
        \caption{This figure exhibits some samples from our dataset. The 2x5 grid presents ten randomly selected high-resolution images (each 3x1024x1024 pixels) without class labels. This offers an overview of the image quality and diversity within the dataset while maintaining the anonymity of the characters depicted. The images showcase various lighting conditions and angles.}
        \label{Figure 2}
    \end{figure}
    are developing a hybrid approach that combines ProGAN's realistic image generation with SAGAN's ability to capture intricate hand gesture details in sign language. Detailed model architecture and dataset descriptions are provided in the following subsections.
    
    \begin{figure}[b]
                \centering
                \includegraphics[width=.488\textwidth, height = 0.25\textwidth]{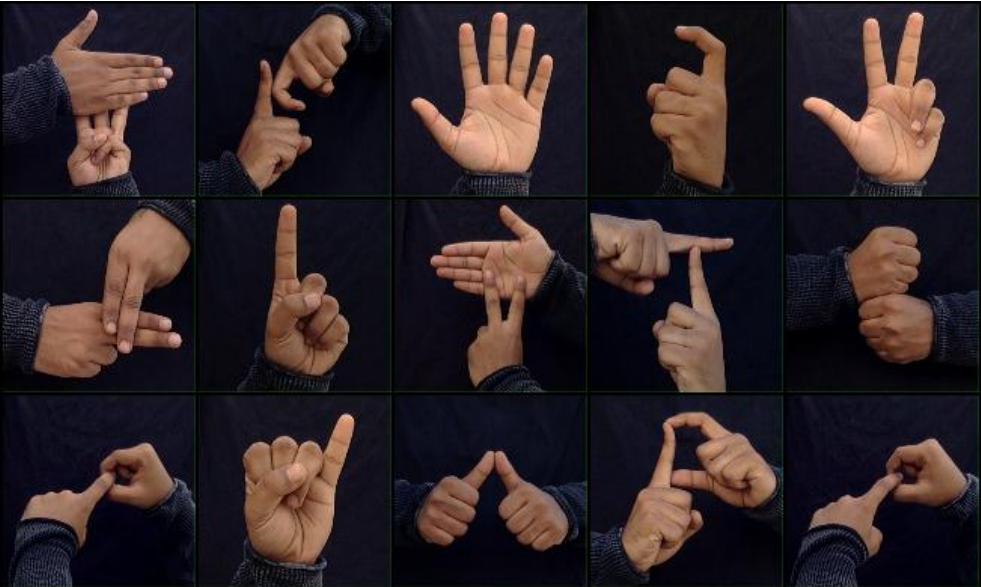}
                \caption{Some random samples from the dataset provided by \cite{b32}. All images in this dataset have dimensions of 3x128x128 and are displayed accordingly.}
                \label{figure 3}
            \end{figure}
 \subsection{Dataset}
    Due to the less availability of high-quality ISL text labeled image datasets, we are introducing our own dataset, which contains high-quality, clean pictures of the English ISL alphabets, numbers, and 129 words. This dataset offers a collection of 2,47,500 high-quality images categorized into 165 unique classes, which contain 26 English alphabet letters, numbers from 0 to 9, and 129 commonly used words, with each class containing more than 1500 images of signs extracted from self-recorded videos, each at a resolution of 3x1024x1024. All sign language images in our dataset are recorded following the standardised images provided by the Indian Sign Language Dataset for Continuous Sign Language Translation and Recognition (ISL-CSLTR) \cite{b38}. The ISL-CSLTR dataset contains five frames for each word and sentence, each given by five different native Indian signers. Our images are captured in real-world backgrounds; no image segmentation is performed to maintain their realism and feature richness. Figure \ref{Figure 2} shows some random samples from it. Performing generation on this dataset is quite challenging because of varying intensity, angles, and complex backgrounds. \\We also tested our model on the ISL alphabet dataset provided by Gunji, Bala Murali, et al.\cite{b32}. This dataset comprises 35 classes with 42,000 images, covering all English letters and numbers from 1 to 9, each containing 1,200 images. This dataset encompasses images with a resolution of 3x128x128, showcasing a variety of lighting conditions and viewing angles. The focus of the dataset is on hand gestures, particularly finger signs. Background elements are removed through segmentation and cropping techniques to isolate the hands for clearer recognition. Figure \ref{figure 3} shows some samples from it.

            
     \subsection{Architecture}
     Our proposed architecture consists of two main components: a generator and a discriminator (critic) network, as illustrated in Figure\ref{Figure 1}. The spatial resolution of the layers in both networks changes across multiple stages. In the generator, each stage's layers double in resolution, starting at 8x8 to capture large-scale structures and increasing to capture finer details. The generator takes two inputs in the first stage: a Gaussian latent noise and a class label. After embedding the class label, it is concatenated with the latent noise, passing through the progressively growing generator networks to produce detailed images. In the generator, each stage includes two weight-standardised convolutional layers to reduce the Lipschitz constant of the loss and the gradient, as recommended by \cite{b39}. Mathematically, this process is defined as follows:
    \begin{equation}
    y = \hat{W} \ast x,
    \end{equation}
    where \(\hat{W} \in \mathbb{R}^{O \times I}\) denotes the standardised weights in the convolution layer and \(\ast\) denotes the convolution operation. For \(\hat{W} \in \mathbb{R}^{O \times I}\), \(O\) is the number of the output channels, and \(I\) is the number of input channels within the kernel region of each output channel. \(\hat{W}\) at point (\(i, j\)) is defined below.
    \begin{equation}
    \hat{W}_{i,j} = \frac{W_{i,j} - \mu_{W_{i,\cdot}}}{\sigma_{W_{i,\cdot}}}
    \end{equation}
    where 
    \[ \mu_{W_{i,\cdot}} = \frac{1}{I} \sum_{j=1}^{I} W_{i,j} \] \[ \sigma_{W_{i,\cdot}} = \sqrt{\frac{1}{I} \sum_{j=1}^{I} W_{i,j}^2 - \mu_{W_{i,\cdot}}^2 + \epsilon} \]
    In many cases, the magnitude of the generator's and discriminator's weights can become uncontrolled due to the adversarial training process. We employed pixel-wise feature vector normalisation in every stage to address this issue. This technique ensures that each feature vector has a unit norm, thereby stabilising the training process of the neural networks. The normalised feature vector \( \hat{V}_{ij} \) at point \( (i,j)\) can be evaluated from the original feature vector \( V_{ij} \) using the following equation:
    \begin{equation}
    \hat{V}_{ij} = \frac{V_{ij}}{\sqrt{\frac{1}{N} \sum_{k=0}^{N-1} (V_{kij})^2 + \epsilon}}
    \end{equation}
    where \( \epsilon = 10^{-8} \), \( N \) is the number of feature maps. Further stages of the network begin by upsampling the feature map, followed by two convolution operations. The convolution layer uses a 3x3 kernel and applies LeakyReLU as the activation function. The final stage of the generator network consists of three convolutional layers. The third convolutional layer uses a 1x1 kernel with a linear activation function to convert the feature map into an RGB image. Within the generator after every stage, a specific layer named toRGB transforms a feature map into 3-channel images of a particular resolution. This layer is applied at the culmination of a progressive stage to obtain the RGB output. However, when the output of one progressive stage proceeds to the next, it bypasses the conversion to RGB using toRGB layer. After up-scaling, feature maps of the previous stage fade up with the current stage using a trainable scale factor $\alpha$. The output image \( I_{\text{out}} \) at \( n \)-th stage is given by the equation:
    \begin{equation}
    I_{\text{out}} = (1 - \alpha) I_{\text{n-1}} + \alpha I_{\text{n}}
    \end{equation}
    where \( I_{\text{out}} \) is the final RGB output at the \( n \)-th stage, \( I_{\text{n-1}} \) is the upsampled RGB output of the previous stage, and \( I_{\text{n}} \) is the RGB output by \( n \)-th stage alone. In general, images start possessing finer detail after a 64x64 resolution. At the 64x64 resolution stage, in between two convolution layers, a Self-attention layer is added. In Figure \ref{Figure 1}, the first Self-attention layer may appear before the 64x64 resolution layer, but it is present between two convolution layers of 64x64. Self-attention allows the convolution layer to focus on specific parts of the image regions that are most relevant in the images with their conditions. This attention layer takes a 256 input feature map of 64x64 dimension and incorporates ReLU as an activation function. This attention layer can focus on significant details in an image at 64x64 resolution. Another Self-attention layer is applied between two convolution layers of resolution 128x128. It processes 128 feature maps of 128x128 resolution and incorporates ReLU non-linearity. Despite being computationally intensive, this attention mechanism significantly enhances the clarity and detail of the output at higher resolution. Subsequently, the output progresses through 256x256 and 512x512 resolution stages, improving its resolution and quality. Finally, the toRGB layer converts it into an RGB image and applies the tanh activation function. 

    The critic network, whose structure is almost a mirror        
    image of the generator network receives the generator's output and the class label while training. It contains an embedding layer that embeds the labels into the total number of classes. The embedded labels then pass through a linear layer, mapping them to a dimension of 512x512. After reshaping to 1x512x512, the embedded label feature maps are concatenated with the corresponding image and processed through the critic network. Critic network layers progressively decrease in resolution by a factor of 2 in many stages. Each progressive stage contains two convolution layers of kernel 3x3 and a layer to down-sample the feature map. The first stage contains a layer fromRGB, which does the opposite of toRGB; it takes embedded four-channel features and converts them into multiple feature maps. The last progressive stage contains a convolution layer which takes a feature map of 512x1x1 dimension and converts it into a 1x1x1 shape. Finally, this output is reshaped into the label. Like the generator, the critic network also has two Self-attention layers. The first attention layer takes 128 feature maps of 128x128 dimension, and the second attention takes 256 feature maps of 64x64 dimension. The final stage of the critic network incorporates a minibatch standard deviation (minibatch stddev) layer, two convolutional layers, and a fully connected layer with linear activation. Minibatch stddev introduces an additional feature map to the discriminator network, capturing the standard deviation of the feature maps across a minibatch of images. This technique helps the discriminator identify whether the generator produces similar images by accounting for variations within the batch, thereby encouraging the generator to create more diverse and realistic images. The fully connected layer then converts the feature map into the corresponding label.
        \begin{figure*}[ht]
            \centering 
            \includegraphics[width=\linewidth, height = 0.25\textwidth]{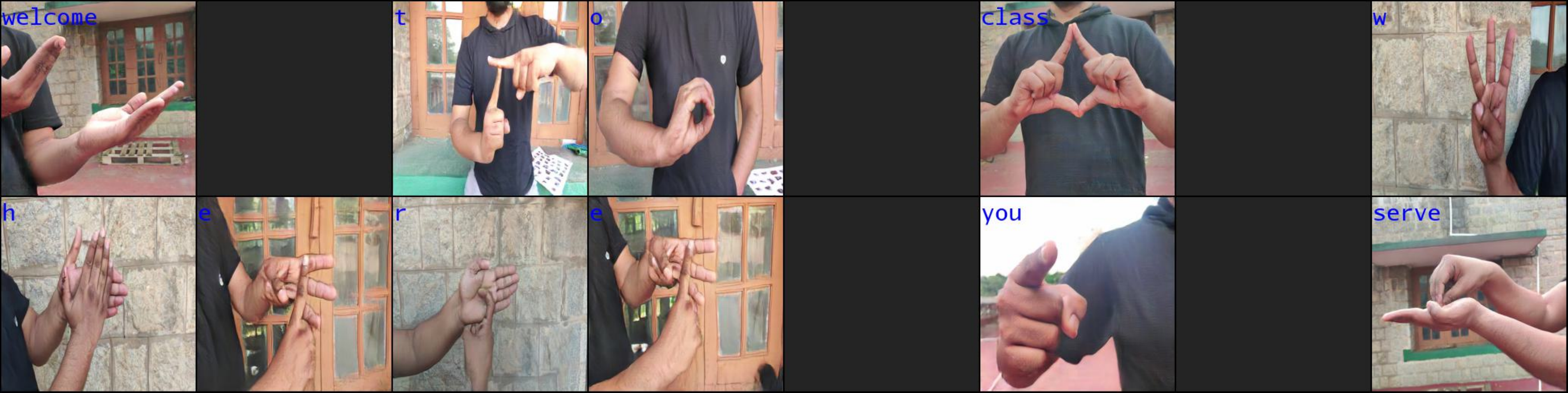}
            \caption{This figure shows the generated output for the sentence, "Welcome to class where you serve," by our model trained on our dataset. A given sentence is divided into constituent words, and a sign language image is generated corresponding to each word. If any word does not belong to a corpus of 129 words, then the model generates outputs for all letters of that word. Complete black images in the output sequence represent space or punctuation in the original sentence. For reference, the corresponding words or letters are displayed in blue colour at the top left corner of each generated image.}
            \label{Figure 4}    
        \end{figure*}
I\subsection{Configurations}
    The complete setup is trained with batch sizes of 32, 32, 32, 32, 16, 16, and 8 for resolutions of 8x8, 16x16, 32x32, 64x64, 128x128, 256x256, and 512x512, respectively. A latent noise vector with dimensions 512x1x1 is fed into the generator each time to produce the output. Generated image values lie between [-1, 1] because of the tanh activation function at the end. In most convolution layers, leaky ReLU non-linearity is used with a leakiness of 0.2. 
    Both networks are trained using the Adam optimiser with a learning rate of $10^{-3}$, $\beta_1=0.0$, and $\beta_2=0.99$. We are using state of the art WGAN-GP loss function. This loss function involves a gradient penalty, which enforces the Lipschitz constraint more effectively by penalising the norm of the gradient of the critic function concerning its input. This approach leads to better convergence, reduced mode collapse, and a more meaningful loss metric, making WGAN-GP a powerful tool for training GANs to generate high-quality, diverse samples. The critic loss with gradient penalty can be defined as:
    \begin{align}
    L_{\text{critic}} = & \ \mathbb{E}_{\tilde{x} \sim \mathbb{P}_g} [D(\tilde{x})] - \mathbb{E}_{x \sim \mathbb{P}_r} [D(x)] \nonumber \\
    & + \lambda \mathbb{E}_{\hat{x} \sim \mathbb{P}_{\hat{x}}} \left[ (\|\nabla_{\hat{x}} D(\hat{x})\|_2 - 1)^2 \right]
    \end{align}
    where:
    \begin{itemize}
        \item \(D\) is the critic function.
        \item \(\mathbb{P}_r\) is the real data distribution.
        \item \(\mathbb{P}_g\) is the model distribution, implicitly defined by \(\tilde{x} = G(z)\), where \(G\) is the generator function and \(z\) is sampled from a prior distribution \(p(z)\).
        \item \(\mathbb{P}_{\hat{x}}\) is defined by uniformly sampling along straight lines between pairs of points sampled from \(\mathbb{P}_r\) and \(\mathbb{P}_g\).
        \item \(\lambda\) is the gradient penalty coefficient. We have used \(\lambda\) = 10.
    \end{itemize}
    The generator loss can be defined as:
    \begin{equation}
    L_{\text{generator}} = -\mathbb{E}_{\tilde{x} \sim \mathbb{P}_g} [D(\tilde{x})]
    \end{equation}
    
\section{Results}
\begin{figure}[t]
            \centering 
            \includegraphics[width=.488\textwidth, height = 0.11\textwidth]{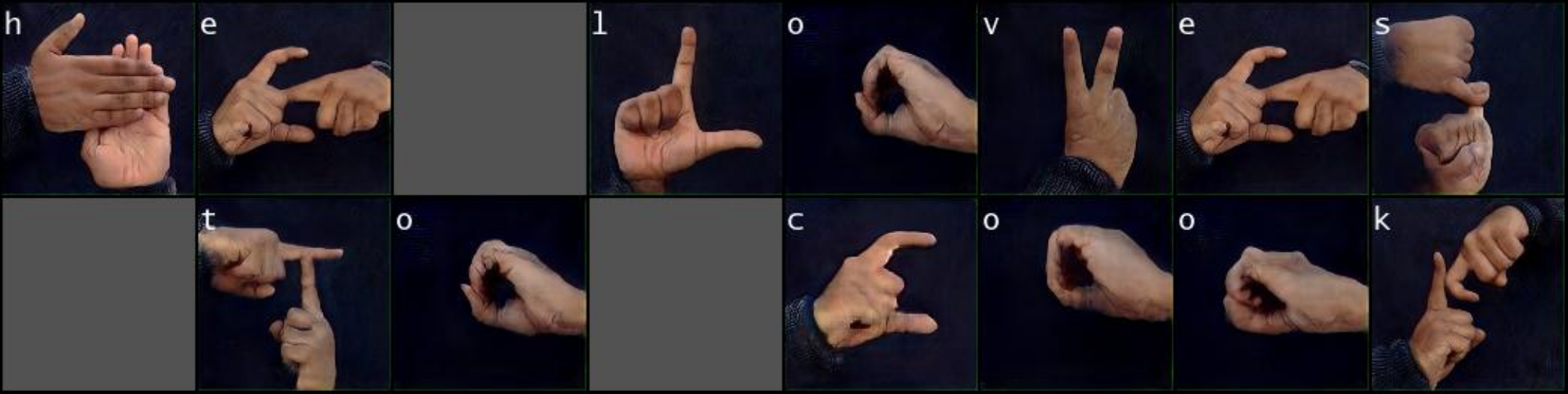}
            \includegraphics[width=.488\textwidth, height = 0.11\textwidth]{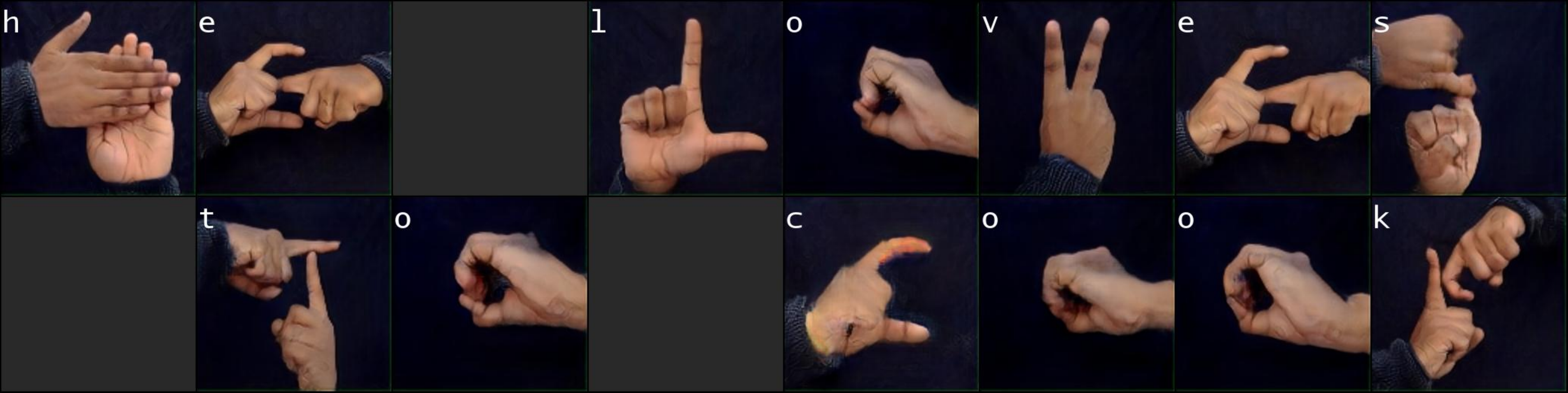}
            \caption{A qualitative comparison of our model with ProGAN output. Both models are trained on \cite{b32} dataset till the same epochs. The first row represents the output of ProGAN, and the last row shows the output from our model.}
            \label{Figure 5}
    \end{figure}

        Figure \ref{Figure 4} illustrates the complete text-to-sign language generation process. Model processes a sentence by breaking it down into individual words. If a word belongs to a predefined set of 129 known classes, it generates an image specific to that class. However, suppose the word isn't recognized within these 129 classes. In that case, the system generates a sequence of images for each letter within the word. Using a well-trained InceptionV3\cite{b20} classifier, we identify the word and letter of the generated images, which are then combined to form words and sentences. The BLEU (1-4) scores corresponding to a paragraph having 139 words after conversion from generated images to text are presented in Table I. For quantitative comparison, we describe the IS \cite{b33} and FID \cite{b34} value on Gunji, Bala, et al.\cite{b32} dataset in Table III, comparing the results of the Conditional Image Synthesis with Auxiliary Classifier GANs (ACGAN)\cite{b40}, and ProGAN model with our model. On this dataset, we observe an improvement of a maximum of 2.47 and 32.12 in IS and FID, respectively. The improvement in IS and FID scores on our dataset are 3.2 and 30.12, respectively. These improvements in metric value suggest that using attention can be highly beneficial. For qualitative comparison, four generated samples from ProGAN and our model are given in Figure \ref{Figure 5}. Seff-attention layers significantly improve the model output, especially when the image has a complex structure. The self-attention mechanism allows many specific feature representations, leading to better outcomes. The artifacts and inconsistency in output are reduced because of attention, 
         \begin{table}[hbt]
        \caption{BLEU score of the models(at 64x64 resolution) using a paragraph having 139 words}
        \begin{center}
            \renewcommand{\arraystretch}{1.2} 
            \begin{tabular}{|p{1.3cm}| p{1.3cm}| p{1.3cm}| p{1.3cm}| p{1.3cm}|}
                \hline
                \textbf{Resolution} &\hspace{2mm} 
                \textbf{\textit{BLEU-1}} &\hspace{2mm} 
                \textbf{\textit{BLEU-2}} &\hspace{2mm} \textbf{\textit{BLEU-3}} &\hspace{2mm} \textbf{\textit{BLEU-4}} \\
                \hline
                \textbf{ProGAN} & 13.9535 & 4.05127 & 0 & 0\\
                \hline
                \textbf{Our model}& \textbf{33.6957} & \textbf{16.0996} & \textbf{8.5298} & \textbf{5.0438}\\
                \hline
            \end{tabular}
            \label{Table I}
        \end{center}
    \end{table}
        making textures and small patterns in images clearer and more distinguishable. We are applying self-attention at two different resolution stages, and Table II contains a comparison of the FID scores after the application of attention at these stages. The attention layer at the 64x64 resolution stage yields an FID score of 29.7, while at the 128x128 resolution stage, it achieves a score of 28.8. When combining both resolutions, the FID score improves to 23.33. From this experimentation, we can conclude that more attention layers can enhance performance.

     \begin{table}[hbt]
        \centering
        \caption{FID scores with the application of the Attention layer at two different resolution stages.}
        \label{tab:fid_scores}
        \renewcommand{\arraystretch}{1.2}
        \begin{tabular}{|p{2cm}|p{1.5cm}|p{1.5cm}| p{1.5cm}|}
        \hline
        \multicolumn{4}{|c|}{\textbf{FID score}} \\
        \hline
        \textbf{Resolution} & 64x64 & 128x128 & At Both \\
        \hline
        \textbf{Score} & 29.701 & 28.803 & 23.331 \\
        \hline
        \end{tabular}
    \end{table} 

            \begin{table*}[hbt]
                \caption{A quantitative comparison of the ACGAN, ProGAN and Our model on \cite{b32} dataset at different resolutions}
                \begin{center}
                    \renewcommand{\arraystretch}{1.2} 
                    \begin{tabular}{|p{1.5cm}|p{1.5cm}|p{1.5cm}|p{1.5cm}|p{1.5cm}|p{1.5cm}|p{1.5cm}|p{1.5cm}|p{1.5cm}|}
                        \hline
                        \multirow{2}{*}{\textbf{Model}} & \multicolumn{4}{|c|}{\textbf{Inception score}} & \multicolumn{4}{|c|}{\textbf{FID score}} \\
                        \cline{2-9} 
                        & \textbf{32x32} & \textbf{64x64} & \textbf{128x128} & \textbf{256x256} & \textbf{32x32} & \textbf{64x64} & \textbf{128x128} & \textbf{256x256} \\
                        \hline
                        \textbf{ACGAN} & -- & 3.91757 & 7.1653 & -- & -- & 223.1019 & 148.8337 & -- \\
                        \hline
                        \textbf{ProGAN} & 1.91 & 9.90086 & 34.8503 & 35.6303 & 96.7992 & 55.4868 & 67.4140 & 67.7980 \\
                        \hline
                        \textbf{Our Model} & \textbf{2.3265} & \textbf{10.6397} & \textbf{35.6718} & \textbf{37.3203} & \textbf{64.6768} & \textbf{30.5914} & \textbf{38.46} & \textbf{39.8212} \\
                        \hline
                    \end{tabular}
                    \label{tab:model_performance}
                \end{center}  
            \end{table*}
         
\section*{Conclusion}
    This work presented an improved generative model designed to create high-quality sign language images. We added the idea of two very popular generative models to get very clear outcomes. Our proposed model leverages attention, Progressive growth of GAN, and state-of-art loss function to achieve 
    exceptional results. Extensive evaluations demonstrated that our model surpasses ProGAN in terms of finger definition, Inception Score, FID score, and spatial structures in images. This improved performance paves the way for real-world applications in sign language education, communication tools, and more. The loss function and the placement of attention layers within the architecture also influence the performance. Creating our high-quality ISL dataset, featuring diverse lighting conditions and a clear background maybe a valuable resource for the research. This dataset can be further utilised in future works to explore advancements in sign language image generation, recognition, and potentially even video generation using diffusion models.

\end{document}